\documentclass{article}

\usepackage[preprint]{nips_2018}

\usepackage{microtype}
\usepackage{booktabs} 

\usepackage{hyperref}
\usepackage{url}
\usepackage{amsbsy}
\usepackage{amsfonts}
\usepackage{amsmath}
\usepackage{graphicx}
\usepackage{subcaption}
\usepackage{xcolor}
\usepackage{comment}
\usepackage{wrapfig}

\definecolor{darkblue}{rgb}{0.0, 0.0, 0.55}
\hypersetup{
    colorlinks=true,
    citecolor=darkblue
}
\def\bx{{\boldsymbol x}}
\def\by{{\boldsymbol y}}
\def\bz{{\boldsymbol z}}

\def\bU{{\boldsymbol U}}

\def\bW{{\boldsymbol W}}
\def\bX{{\boldsymbol X}}
\def\bY{{\boldsymbol Y}}

\def\bG{{\boldsymbol G}}

\def\bL{{\boldsymbol L}}

\def\bl{{\boldsymbol l}}

\def\R{{\mathbb{R}}}
\newcommand{\conv}[1]{$\left[\begin{array}{ll} \text{3}\times \text{3} \text{ conv} \end{array}\right] \times \text{#1}$}
\newcommand{\gconv}[2]{$\left[\begin{array}{ll} \text{#1} \text{-conv} \end{array}\right] \times \text{#2}$}

\title{Learning Depthwise Separable Graph Convolution \\ from Data Manifold}

\author{
  Guokun Lai \\
  Language Technologies Institute\\
  Carnegie Mellon University\\
  \texttt{guokun@cs.cmu.edu} \\
  \And
  Hanxiao Liu \\
  Language Technologies Institute\\
  Carnegie Mellon University\\
  \texttt{hanxiaol@cs.cmu.edu} \\  
  \And
  Yiming Yang \\
  Language Technologies Institute\\
  Carnegie Mellon University\\
  \texttt{yiming@cs.cmu.edu} \\  
}

\begin{document}

\maketitle

\begin{abstract}
Convolution Neural Network (CNN) has gained tremendous success in computer vision tasks with its outstanding ability to capture the local latent features. Recently, there has been an increasing interest in extending convolution operations to the non-Euclidean geometry. Although various types of convolution operations have been proposed for graphs or manifolds, their connections with traditional convolution over grid-structured data are not well-understood. In this paper, we show that depthwise separable convolution can be successfully generalized for the unification of both graph-based and grid-based convolution methods. Based on this insight we propose a novel Depthwise Separable Graph Convolution (DSGC) approach which is compatible with the tradition convolution network and subsumes existing convolution methods as special cases. It is equipped with the combined strengths in model expressiveness, compatibility (relatively small number of parameters), modularity and computational efficiency in training. Extensive experiments show the outstanding performance of DSGC in comparison with strong baselines on multi-domain benchmark datasets.
\end{abstract}

\section{Introduction}

Convolution Neural Network (CNN) \citep{lecun1995convolutional}, also referred to as 2D-convolution in the following paper, has been proven to be an efficient model family in extracting hierarchical local patterns from grid-structured data,
which has significantly advanced the state-of-the-art performance of a wide range of machine learning tasks, including image classification, object detection and audio recognition \citep{lecun2015deep}. Recently, growing attention has been paid to dealing with data with a non-grid structure, such as prediction tasks in sensor networks \citep{xingjian2015convolutional}, transportation systems \citep{li2017graph}, and 3D shape correspondence application in the computation graphics \citep{bronstein2017geometric}. How to replicate the success of CNNs for manifold-structured data remains an open challenge.

In this paper, we provide a unified view of the 2D-convolution methods and the graph convolution (including the geometric convolution) with the label propagation process \citep{zhu2003semi}. To best of our knowledge, it is the first time that the 2D-convolution and graph convolution proposed in \citep{kipf2016semi} are unified mathematically. It helps us better understand and compare the difference between them, and shows that the fundamental difference can be summarized as two points, (1) 2D-convolution learns spatial filters from the data. (2) Spatial filters in 2D-convolution are channel-specific.

Many graph convolution and geometric convolution methods have been proposed recently. The spectral convolution methods \citep{bruna2013spectral,defferrard2016convolutional,kipf2016semi} are the mainstream algorithms developed as the graph convolution methods. Their theory is based on the graph Fourier analysis \citep{shuman2013emerging}. Another group of approaches are geometric convolution methods, which focus on various ways to leverage spatial information about nodes \citep{masci2015geodesic,boscaini2016learning,monti2016geometric,gilmer2017neural}. Existing models mentioned above are either fully trusting the given graph or applying one graph filter across all channels, which are corresponding to the two differences between the traditional 2D-convolution and the graph convolution. Firstly, as a result of trusting the given graph, namely only using a given graph filter across the whole model, the model ability to discover the special graph filters from the supervision data is limited. And applying one graph filter across all channels would also introduce several drawbacks to the model as follows. (1) It makes the mathematical formulation of the graph convolution methods incompatible with the traditional 2D-convolution. (2) The model cannot propagate information with different diffusion patterns in one layer. (3) The image recognition experiment in Section \ref{sec:cifar} shows that multiple filters are critical to the model performance in the task which requires extracting complex local features. Some models, such as MoNet \citep{monti2016geometric} and Graph Attention Network \citep{velivckovic2017graph}, try to model multiple filters by simultaneously learning $K$ sub-layers, each with one global filter, and summarizes them as one layer. However, this approach would lead to a larger number of parameters and more expensive computation cost, and it is still incompatible with the traditional 2D-convolution method.    

In this paper, we derive a novel graph convolution approach directly from the 2D-convolution method. We propose the Depthwise Separable Graph Convolution (DSGC), which inherits the strength of depthwise separable convolution that has been extensively used in different state-of-the-art image classification frameworks including Inception Network \citep{szegedy2016rethinking}, Xception Network \citep{chollet2016xception} and MobileNet \citep{howard2017mobilenets}. Compared with previous graph and geometric methods, the DSGC is more expressive and compatible with the depthwise separable convolution network, and shares the desirable characteristic of small parameter size as in the depthwise separable convolution. In experiments section, we evaluate the DSGC and baselines in three different machine learning tasks. The experiment results show that the performance of the proposed method is close to the standard convolution network in the image classification task on CIFAR dataset. And it outperforms previous graph convolution and geometric convolution methods in all tasks. Furthermore, we demonstrate that the proposed method can easily leverage the state-of-the-art architectures developed for image classification to enhance the model performance, such as the Inception module \citep{szegedy2016rethinking}, the dense block \citep{huang2016densely} and the Squeeze-and-Excitation block \citep{hu2017squeeze}.

The main contribution of this paper is threefold:
\begin{itemize}
\item A unified view of traditional 2D-convolution and graph convolution methods by introducing depthwise separable convolution. 
\item A novel Depthwise Separable Graph Convolution (DSGC) for data residing on arbitrary manifolds. 
\item We demonstrate the efficiency of the DSGC module with extensive experiments and show that it can be plugged into existing state-of-the-art CNN architectures to improve the performance for graph tasks. 
\end{itemize}

\section{A Graph Perspective of Convolution}
\label{sec:proposed}
In this section, we provide a unified view of several convolution operations
by showing that they are different message aggregation protocols over the graphs or manifolds.
Unless otherwise specified,
we denote a matrix by $\bX$,
the $i$-th row in the matrix by $\bx_i$,
and the $(i,j)$-th element in the matrix by $x_{ij}$.
Superscripts are used to distinguish different matrices when necessary.
All the operations below can be viewed as a function that transforms input feature maps
$\bX \in \R^{N \times P} $ to output feature maps $\bY \in \R^{N \times Q}$, 
where $N$ is the number of nodes and $P, Q$ are the number of its associated input and output features (channels) respectively.
We use $\bG$ to denote the adjacency matrix of a graph,
and $G(i)$ to denote the set of neighbors for node $i$.
For tasks over sensor networks \citep{xingjian2015convolutional}, transportation graphs \citep{li2017graph} or computational graphics \citep{bronstein2017geometric},
graph $G$ often corresponds to the latent structure of the underlying manifold, and is induced from the spatial coordinates of input data.

\subsection{Convolution over Graphs}
For the operations discussed below,
the filter weights are fully determined by the given graph $G$.
% the filter weights are defined as its normalized adjacency matrix $\bW$.

\subsubsection{Label Propagation}
Label propagation (LP) \citep{zhu2003semi} can be viewed as a simplistic convolution operation to aggregate local information over a graph:
\begin{equation}
\by_{i} = \sum_{j \in G(i)} G_{ij} \bx_{j}
\label{eq:labpro}
\end{equation}
In other words, the feature map for each node is updated as the weighted combination of its neighbors' feature maps. In this case, the numbers of input and output channels are identical.
% \begin{equation}
% y_{iq} = \sum_{j \in \mathcal{N}(i)} w_{ij} x_{jq}
% \label{eq:labpro}
% \end{equation}

% where $X_i$ and $Y_i$ are the latent vector representations of node $i$ before and after the propagation, respectively.

\subsubsection{Graph Convolution}
Graph convolution \citep{kipf2016semi} (GC)
can be viewed as an extension of LP, formulated as:
\begin{equation}
\begin{aligned}
\by_i = \sum_{j \in G(i)} G_{ij} \bz_j \quad where \quad \bz_j = \bx_j \bU \\
\end{aligned}
\label{eq:gcn}
\end{equation}
While both LP and GC utilize the graph structure in $\bG$,
GC has an learnable linear transformation $\bU \in \R^{P \times Q}$ that maps $\bx_j$ into the intermediate representation $\bz_j$.
This additional step enables GC to capture the dependencies among channels.

%$\bU \in \R^{P \times Q}$ represents a linear transformation.
%Following the notation in DSC, $\bW$ is still named as the spatial filter and $\bU$ is named as the channel filter.

\subsection{Convolution over Grid-Structures}

Here, we write the convolution methods over 2D-grid in the Label Propagation framework. Let $\Delta_{ij}$ be the coordinate offset from $i$-th node to $j$-th node, we say $j \in G(i)$ if $j$ is one of $i$'s $k$-nearest neighbors based on the relative distance $|\Delta_{ij}|$.

\subsubsection{Full Convolution}
The full convolution \citep{lecun1995convolutional} can be formulated as

\begin{equation}
y_{iq} = \sum_{j \in G(i)} \sum_{p=1}^P w^{(pq)}_{\Delta_{ij}} x_{jp}
\label{eq:full}
\end{equation}

For Euclidean grid-structured data such as images,
$\Delta_{ij}$ denotes the offset between pixel $i$ and pixel $j$,
and $G(i)$ contains pixel $i$'s surrounding pixels.
For example, the size of $G(i)$, or $k$, is $9$ for $3 \times 3$ convolution and 25 for $5\times 5$ convolution,
corresponding to the size of the receptive field.
The full convolution operation captures the channel correlation and spatial correlation simultaneously by $\bW^{(pq)}$.

% \subsubsection{Depthwise Convolution}
% \begin{align}
% y_{iq} &= \sum_{j \in \mathcal{N}(i)} w^{(q)}_{\Delta_{ij}} x_{jq}
% \end{align}

\subsubsection{Depthwise Separable Convolution}

%A representative example of convolution operation for 2D-grids is depthwise separable convolution (DSC) \citep{chollet2016xception}, which can be viewed as a factorized version of full convolution \citep{lecun1995convolutional} under the intuition that the channel correlation and spatial correlation can be decoupled.
The Depthwise Separable Convolution (DSC) \citep{chollet2016xception} is a factorized version of full convolution under the intuition that the channel correlation and spatial correlation can be decoupled.
In practice, DSC is able to achieve comparable performance as full convolution with a substantially smaller number of parameters.
We focus on DSC here due to its simplicity and intimate connections to Graph Convolution.
DSC can be formulated in a graph-based fashion as follows
\begin{align}
y_{iq} &= \sum_{j \in G(i)} w^{(q)}_{\Delta_{ij}} z_{jq} \quad where \quad \bz_j = \bx_j \bU 
\label{eq:sepa}
\end{align}

The formulation of DSC is analogous to GC by substituting $\bG$ in eq. \eqref{eq:gcn} with $\bW$. 
However, unlike LP and GC which directly utilize the graph $\bG$ to define their filter weights, weights $\bW$ in eq. \eqref{eq:sepa} is a learnable lookup table of size $Q \times R$,
where $R$ is the number of possible choices for $\Delta_{ij}$.
For example, $R=9$ for $3 \times 3$ convolution, since $\Delta_{ij}$ can take any value in $\{-1, 0, 1\} \times \{-1, 0, 1\}$.

\section{Depthwise Separable Graph Convolution}
\label{sec:model}

\subsection{Motivation}
%Although DSC is not proposed for arbitrary manifolds,
We notice that DSC is more powerful than GC in the following aspects:
\begin{enumerate}
\item The spatial filter in GC is fully determined once the graph is given \footnote{The linear transformation $\bU$ in GC is not a graph/spatial filter,
as it only fuses the information across the channels.}, but the spatial filters in DSC are learned automatically from data.
This means GC would fully trust the given graph even if it is suboptimal for the task and data on hand.

\item Compared with full convolution and DSC, which are capable of modeling channel-specific convolution filters, GC uses a global spatial filter for all channels (features), which can be viewed as a restricted version of DSC. Thus a GC module is unable to simultaneously capture or fuse diverse information based on different channels/features over the graph.
% \yiming{Hanxiao: Check the above sentence.}
%the diverse effect of different graph filters corresponding to different information diffusion patterns over the graph.
\end{enumerate}
On the other hand, while GC is generally applicable to arbitrary graphs, the DSC method so far is only designed for regular grid-based structures and hence only applicable to the domains like image processing,
where the pixels naturally form a grid structure.
For nodes scattered with arbitrarily spatial coordinates, the number of possible choices for $\Delta_{ij}$ can be infinite.  That is, using a lookup table $\bW$ to memorize the filter weights for each $\Delta_{ij}$ is no longer feasible. This makes traditional DSC not directly applicable to arbitrary graphs.
% \yiming{Hanxiao: Is the last sentence (mine) correct?}
% On the other hand, GC suffers from the restriction that all channels have to share the same given spatial filter. 

\subsection{Proposed Method}
\label{sec:graphconv}
To address the aforementioned limitations of Graph Convolution,
we propose Depthwise Separable Graph Convolution (DSGC),
which naturally generalizes DSC and GC as:
\begin{align}
y_{iq} &= \sum_{j \in G(i)} w^{(q)}(\Delta_{ij}) z_{jq} \quad where \quad \bz_j = \bx_j\bU 
\label{eq:proposed}
\end{align}
where we slightly abuse the notation by overloading $w^{(q)}(\cdot)$ as a function (neural network) that maps $\Delta_{ij}$ to a real scalar, namely the predicted filter weight for the $q$-th channel.

The key distinctions in our formulation are as follows.
\begin{enumerate}
\item Different from DSC (eq.\ \eqref{eq:sepa}),
the filter weight is calculated using a ``soft'' function approximator.  That is,
DSGC predicts the convolution filter weights from $\Delta_{ij}$ via the function
instead of memorizing them in a look-up table.
In our experiments, function $w^{(q)}(\cdot)$ is implemented as a two-layer MLP.
\item Different from GC (eq.\ \eqref{eq:gcn}), DSGC enables the learning of channel-specific spatial convolution filters
(channels are indexed by $q$ in eq.\ \eqref{eq:proposed}).
% \yiming{Hanxiao: Do you mean "q" is the index for different channels? It is not even defined in formular 4.}
This amounts to simultaneously constructing channel-specific graphs under the different node-node similarity metrics, where the metrics are implicitly defined by neural networks and hence, are jointly optimized during the training.
\end{enumerate}

The idea of predicting the filter weights has also been explored in Message Passing Neural Network (MPNN) \citep{gilmer2017neural}. However, MPNN learns only a global function across all channels, hence, MPNN is incapable of capturing channel-specific spatial filters as in DSC.

\subsection{Parameter Grouping Strategy}
Overfitting is a common issue in graph-based applications
due to limited data available. 
To alleviate this issue,
a simple strategy is to group the original $Q$ channels into $C$ groups,
where $D = Q/C$ channels in the same group would share the same filter:
%This amounts to sharing the weights in convolution neural networks,
\begin{equation}
w^{(q)}(\cdot) = w^{(q')}(\cdot) \quad if \quad \lfloor \frac{q}{D} \rfloor = \lfloor \frac{q'}{D} \rfloor
\label{eq:DSGC}
\end{equation}
where $\lfloor \cdot \rfloor$ denotes the floor function.

\subsection{Filter Normalization}

\label{sec:normalization}

% The context of each node in any given generic graph,
% namely its connection pattern with neighbors,
% can be non-stationary over different parts of the graph, while it is constant in the 2d-grid graphs.
A common practice in label propagation and graph convolution is to normalize the adjacency matrix for $G$.
In DSGC,
a natural way to carry out normalization
is to apply a softmax function as the final layer of the filter weights predictor, to ensure that $\sum_{j\in G(i)} w^{(q)}(\Delta_{ij}) = 1$
for each $i$.
% which can be written as $\tilde{\bw}_i = softmax(\bw_i)$,
% where $\bw_i$ stands for the $i$-th row of spatial filter $\bW$ learned by a neural network.
We empirically found that normalization leads to improved performance and faster convergence. 

% In the following experiments, we use the proposed depthwise separable graph convolution with a linear highway bypass as the basic convolution component and imitate the rest setting of the standard convolution neural network to solve different machine learning tasks.

\section{Closely Related Models}
\label{sec:related}

Several representative works in graph convolution are worth discussing w.r.t. their connections to ours.
\begin{comment}
\subsubsection{Full Convolution}
The early work can be traced back to the full convolution \citep{lecun1995convolutional} in the form of:

\begin{equation}
y_{iq} = \sum_{j \in G(i)} \sum_{p=1}^P w^{(pq)}_{\Delta_{ij}} x_{jp}
\end{equation}

Different from our DSGC (Eq.\ref{eq:sepa}), it captures the channel correlation and spatial correlation simultaneously by $\bW^{(pq)}$, which leads to a larger number of parameters compared to DSC-based method including DSGC.
\end{comment}

\subsection{Spectral Convolution Methods}
The Spectral Network \citep{bruna2013spectral}
is derived from the graph signal processing work \citep{shuman2013emerging}, which generalizes Fourier analysis for its use in the graph domain as:
\begin{equation}
\begin{aligned}
y_{iq} &= \sum_{j \in G(i)} \sum_{p=1}^P w^{(pq)}_{ij} x_{jp}  \\
\quad where  & \quad \bW^{(pq)} = \Phi \Lambda^{(pq)} \Phi^T
\label{eq:spectral}
\end{aligned}
\end{equation}
where $\Phi \in \R^{n \times n}$ are the eigenvectors of $G$'s graph Laplacian matrix, and $\Lambda$ are learnable nonparametric filters from the training data. The Spectral Network can be matched with the full convolution (eq.\eqref{eq:full}), but with the different filter subspace, in other words, with different basic filters.
Limitations of Spectral Networks include its high computation cost due to eigen-decomposition of the graph Laplacian, the lack of spatial locality and the large number of parameters which grows linearly over the graph size.

These limitations are partially addressed in the Chebyshev Networks \citep{defferrard2016convolutional} (ChebyNet),
which approximates the non-parametric filters as:
\begin{equation}
y_{iq} = \sum_{j \in G(i)} \sum_{k=1}^{K} T_{k}(L)_{ij} z^{(k)}_{iq}, \quad \bz^{(k)}_{i} = \bx_{j} \bU^{(k)}  \label{eq:cheby} \\
\end{equation}
where $T_k(L)$ is the $k$-th order Chebyshev polynomial term.
While being faster than Spectral Networks,
ChebyNet suffers from insufficient expressiveness, similar to the limitations of GC.
The expressiveness of ChebyNet can be improved by enlarging $K$,
which requires a much larger number of parameters and eventually converges to Spectral Networks.

\subsection{Geometric Convolution Methods}
Several geometric convolution methods \citep{masci2015geodesic,boscaini2016learning,monti2016geometric}
are proposed for manifold structured data,
among which MoNet \citep{monti2016geometric} is the state-of-the-art.

The updating formula for MoNet can be written as
\begin{equation}
\begin{aligned}
y_{iq} &= \sum_{j \in G(i)} \sum_{k=1}^{K} w_k(v(i,j)) z^{(k)}_{jq}, \quad \bz^{(k)}_{j} = \bx_j \bU^{(k)} \label{eq:monet} \\
w_k(v) &= \exp\left(-\frac{1}{2}(v - \mu_k)^T \Sigma^{-1}_k(v - \mu_k)\right)
\end{aligned}
\end{equation}
where $v(i,j)$ is the embedding of a node pair similar to $\Delta_{ij}$ in our model,
and $\mu_k, \Sigma_k$ are both learnable parameters.
%and $\Sigma_k$ is constrained as the diagonal matrix.
MoNet can be viewed as an extension of ChebyNet where the graph filters are learned from data instead of being fully determined by a given graph.  However, a graph filter in MoNet is still applied across all channels. In order to have $k$ filters in MoNet, it needs to learn $k$ different channel filters $\bU$. Then the total number of model parameters will grow linearly with $k$. While DSGC only learns a channel filter $\bU$ for one layer. By taking advantage of that, the number of parameters in DSGC would not be significantly growing with $k$ as MoNet. Similar to it, the recently proposed Graph Attention Network (GAT) \citep{velivckovic2017graph} is also required to learn multiple channel filters in order to model multiple filters in one layer, which would lead to a larger number of the model parameters. Furthermore, in the setting that nodes in graph have geometric information, GAT can be viewed as a MoNet extension with the filter normalization trick. We empirically found that the extension exhibits obvious improvement over the original MoNet. In the following parts, we denote it as MoNet with GAT.

Message Passing Neural Networks (MPNN) \citep{gilmer2017neural, schutt2017schnet} are developed for modeling information propagation over graphs, specialized for the prediction tasks in quantum chemistry.
Similar to DSGC, MPNN utilizes a neural network to predict the filter weights for the convolution operations.
The key difference is that while DSGC allows channel-specific graph convolution filters (hence allowing a variety of diffusion patterns over the graph), MPNN learns only a single graph filter function for all channels in a layer. So MPNN can be viewed as a special case of DSGC with $C = 1$ in eq. \eqref{eq:DSGC}. In our experiments, our method consistently outperforms MPNN across tasks in a variety of domains.

\section{Experiments}
\label{sec:experiment}

\subsection{Experimental Design}

Our experiments for evaluating the proposed DSGC approach consists of three parts.  Firstly, we evaluate DSGC on a popular image classification dataset (Sec.  \ref{sec:cifar}).  The purpose is to confirm the strong performance of DSGC in handling grid-based convolution although it is designed for more general graph structures.  Secondly, we compare DSGC and strong methods in the tasks of time series forecasting (Sec. \ref{sec:time}) and text categorization (Sec. \ref{sec:20news}) where grid-based convolutions are invalid but graph convolution would have advantages instead as they are designed for more flexible graph structures. 
Thirdly, we examine the flexibility and effectiveness of using DSGC as a building block (module) in multiple well-known neural network architectures, including Inception \citep{szegedy2016rethinking}, DenseNet framework \citep{huang2016densely} and etc.  (See Appendix \ref{sec:advance}).

For controlled experiments, all the graph convolution methods share the same empirical settings unless otherwise specified, including network structures, the dimension of latent factors, and hyper-parameter tunning process. The neural network used to model the spatial convolution filter ($w^{(q)}(\cdot)$) in eq. \eqref{eq:proposed} is a two-layer MLP with 256 hidden dimensions and tanh activation function. We have conducted ablation tests with the two-layer MLP by changing the number of layers and activation function of each hidden layer, and by trying several weight sharing strategies. The results are very similar; the two-layer MLP provides a reasonable performance with the shortest running time. Appendix \ref{sec:appendix-detail} contains more details, such as the network architecture and model hyper-parameters.
The algorithms are implemented in PyTorch; all the data and the code including baselines are made publicly accessible \footnote{Code: https://github.com/laiguokun/DSGC \\ Data: https://drive.google.com/drive/folders/0BweQMXBkrHAcSkpkejFsOXNId2s?usp=sharing}.

\subsection{Evaluation on Image Classification}
\label{sec:cifar}

\begin{figure}[!th]
\centering
\begin{subfigure}{.25\textwidth}
  \includegraphics[width=\linewidth]{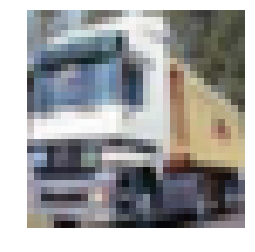}
  \caption{}
\end{subfigure}
\begin{subfigure}{.25\textwidth}
  \includegraphics[width=\linewidth]{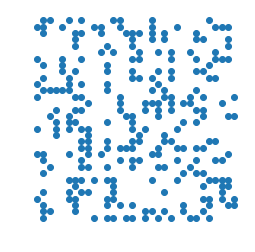}
  \caption{}
\end{subfigure}
\begin{subfigure}{.25\textwidth}
  \includegraphics[width=\linewidth]{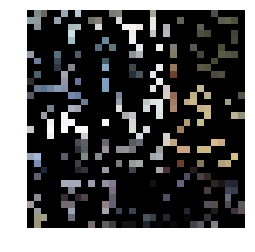}
  \caption{}
\end{subfigure}
\caption{How to construct subsampled CIFAR datasets: (a) is an example image from CIFAR dataset. (b) is the subsampled pixels map. The blue points indicate which points are sampled. (c) is the image after sampling, where the black points are those not being sampled.}
\label{fig:cifar}
\end{figure}

We conduct experiments on CIFAR10 and CIFAR100  \citep{krizhevsky2009learning}, which are popular benchmark datasets in image classification. Both sets contain 60000 images with $32 \times 32$ pixels but CIFAR10 has 10 category labels and CIFAR100 has 100 category labels. Each image is typically treated as a $32 \times 32$ grid structure for standard image-based convolution. To enable the comparison on generic graphs, we create the modified versions of CIFAR10 and CIFAR100 respectively, by subsampling only 25\% of the pixels from each graph. As illustrated in Figure \ref{fig:cifar}, the subsampling result is irregularly scattered nodes for each image. The detailed experiment settings are included in Appendix \ref{sec:appendix-cifar}.

\begin{comment}
As the baselines for comparison we include both the traditional 2D-convolution methods and graph convolution networks as listed: 
(1) standard CNN; 
(2) Xception network \citep{chollet2016xception} which uses the depthwise separable convolution; 
(3) DCNN \citep{atwood2016diffusion}, the method using multi-hops random walk as the graph filters; 
(4) ChebyNet \citep{defferrard2016convolutional}, the method using Chebyshev polynomial to approximate the Fourier transformation of (irregular) graphs; 
(5) GCN \citep{kipf2016semi} which is described in Section \ref{sec:proposed}; 
(6) MoNet \citep{monti2016geometric}, the method using Gaussian function to define the propagation weights over (irregular) graphs.
(7) MPNN \citep{gilmer2017neural}, the method using a global edge network to define the propagation weights.
\end{comment}

For all methods, we use the VGG-13 architecture \citep{simonyan2014very} as the basic framework, and replace its convolution layers with different convolution modules. The experiment results are summarized in Table \ref{tab:cifar}. The best performances among the graph-based neural networks are in bold. Firstly, we observe that Xception and CNN have the best results; this is not surprising because both methods use grid-based convolution which is naturally suitable for image recognition. Secondly, DSGC outperforms all the other graph-based convolution methods, and its performance is very close to that of the grid-based convolution methods. We also see that the models that learn multiple filters (DSGC and MoNet) have better performance than the models that only learn one global graph filter, such as MPNN and GCN. It demonstrates that it is necessary to enable multiple filters in this task. Furthermore, contributed by the depthwise separable convolution and graph sharing technique, our model can achieve a competitive performance without increasing the number of parameters as GCN, the one with the smallest number of parameters among graph convolution approaches. On the contrary, MoNet and ChebyNet have a relatively larger number of parameters in order to model multiple filters in one layer. In Appendix \ref{sec:traintime}, we analyze the computation cost and training time of the proposed DSGC and baseline methods.

\begin{table*}[!ht]
\resizebox{\textwidth}{!}{
\centering
\begin{tabular}{l|c|c|c|c|c|c}
\hline
   &\multicolumn{3}{|c|}{Subsampled Images} & \multicolumn{3}{|c}{Original Images} \\
\hline
Method          & CIFAR10 & CIFAR100 & \#params & CIFAR10 & CIFAR100 & \#params \\
\hline
DCNN \citep{atwood2016diffusion}             
				 & 43.68\% & 76.65\% &12M & 55.56\% & 84.16\% & 50M  \\
ChebyNet \citep{defferrard2016convolutional} 
                 & 25.04\% & 49.44\% &10M & 12.99\% & 36.96\% &19M \\
GCN \citep{kipf2016semi}
				 & 26.78\% & 51.30\% &5.6M & 19.09\% & 41.64 \% &9.8M \\
MoNet (with GAT) \citep{monti2016geometric}          
				 & 21.20\% & 47.87\% &11M & 8.34\% & 29.56\% &20M \\
MPNN \citep{gilmer2017neural}
				 & 22.71\% & 49.03\% &5.6M & 11.01\% & 32.95\%& 9.9M\\      
DSGC (ours)            & \textbf{18.72}\% & \textbf{44.33}\% &5.7M & \textbf{7.31}\% & \textbf{27.29}\% &9.9M \\
\hline
%CNN(Wo mask)     & 80.67\% & 18.3M & -       & -     \\
CNN (VGG-13) \citep{simonyan2014very}              
				 & 18.03\% & 43.42\% &18M & 6.86\% & 26.86\% &18M \\
CNN (Xception) \citep{chollet2016xception}
				 & 17.07\% & 41.54\% & 3.1M & 7.08\% &  26.84\%& 3.1M\\
\hline
\end{tabular}
}
\caption{Test-set error rates on CIFAR10 and CIFAR100. DSGC has the best performance among the graph-based convolution method group (first six methods), and is comparable to the state-of-the-art grid-based convolution methods (VGG-13 and Xception) which are tailed for image classification.}
\label{tab:cifar}
\end{table*}

\subsection{Evaluation on Time Series Forecasting}

\label{sec:time}

In time-series forecasting, we are usually interested in how to effectively utilize the geometric information about sensor networks. For example, how to incorporate the longitudes/latitudes of sensors w.r.t. temporal cloud movement is a challenge in spatiotemporal modeling for predicting the energy output of solar energy farms in the United States. Appendix \ref{sec:appendix-time} provides the formal definition of this task.

We choose three publicly available benchmark datasets for this task:  
\begin{itemize}
\item The U.S Historical Climatology Network (USHCN)\footnote{\url{http://cdiac.ornl.gov/epubs/ndp/ushcn/daily_doc.html}} dataset, used in \cite{bahadori2014fast}, contains daily climatological data from 1,218 meteorology sensors over the years from 1915 to 2000. The sequence length is 32,507. It includes five subsets, and each has a climate variable: (1) maximum temperature, (2) minimum temperature, (3) precipitation, (4) snowfall and (5) snow depth. We use daily maximum temperature data and precipitation data, and refer them as the \textbf{TMAX} and \textbf{PRCP} sets, respectively.
\item The solar power production records in the year of 2006 has the data with the production rate of every 10 minutes from 1,082 solar power stations in the west of the U.S. The sequence length is 52,560. We refer this set of data as \textbf{Solar}.
\end{itemize}
All the datasets have been split into the training set (60\%), the validation set (20\%) and the test set (20\%) in chronological order. 

%Except for the graph convolution methods,  we also add in traditional methods of time series forecasting for comparison, such as (1) Autoregressive model (AR) which predicts future signal using a window of historical data based on a linear assumption about temporal dependencies, (2) Vector autoregressive model (VAR) which extends AR to the multivariate version, namely, the input is the signals from all sensors in the history window, and (3) the LSTNet deep neural network model \citep{lai2017modeling} which combines the strengths of CNN, RNN and AR.  None of those methods is capable of leveraging locational dependencies via graph convolution. We exclude the CNN and Xception methods, the 2D-grid based convolution, which could not be generalized to irregular graphs which we focus on here.

Table \ref{tab:time} summarizes the evaluation results of all the methods, where the performance is measured using the Root Square Mean Error (RMSE). The best result on each dataset is highlighted in boldface. 
%The group of the first three methods does not leverage the spatial or locational information in data. The second group (graph-based convolution methods) consists of the neural network models which leverage the spatial information about sensor networks.  The methods in the second group clearly outperform the methods in the first one, which does not explicitly model the spacial correlation within sensor networks. 
Overall, our proposed method (DSGC) has the best performance on all the datasets, demonstrating its strength in capturing informative local propagation patterns both temporally and spatially. In Appendix \ref{sec:appendix-time}, we include more comparisons with the pure temporal methods. 

%\begin{wraptable}{r}{8.0cm}
\begin{table}[!ht]
\centering
\begin{tabular}{l|c|c|c}
\hline
Method           	& TMAX 	   	 & PRCP 	  & Solar  \\
\hline
%AR                	& 8.2354     & 30.3825    & 0.03195        \\
%VAR               	& 17.9743    & 29.2597    & 0.03296         \\
%LSTNet %\citep{lai2017modeling}          	
%					& 10.1973    & 29.0624    & 0.02865   \\
%\hline
DCNN %\citep{atwood2016diffusion}
					& 6.5188	 & 29.0424	  & 0.02652	  \\
ChebyNet %\citep{defferrard2016convolutional}
					& 5.5823     & 27.1298    & 0.02531     \\
GCN %\citep{kipf2016semi}
					& 5.4671     & 27.1172    & 0.02512     \\
MoNet (with GAT) %\citep{monti2016geometric}
					& 5.8263     & 26.8076    & 0.02564    \\ 
MPNN 				& 5.3331     & 26.4766	  & 0.02496\\
\hline
DSGC (ours)               & $\textbf{5.1438} (\pm 0.0498)$ & $\textbf{25.8228} (\pm 0.249)$ & $\textbf{0.02453} (\pm 0.00022)$\\
\hline
\end{tabular}
\caption{Test-set performance for graph convolution methods on time series prediction tasks measuring in RMSE. For our method, we report the standard deviation of the performance by running the model with 10 random seeds.}
\label{tab:time}
\vspace{-0.5cm}
\end{table}
%\end{wraptable}

\subsection{Evaluation on Document Categorization}
\label{sec:20news}

\begin{wraptable}{r}{6.0cm}
%\begin{table}[!ht]
\centering
\begin{tabular}{l|c}
\hline
Method & Accuracy \\
\hline
Linear SVM$^\dagger$       & 65.90\% \\
Multinomial NB$^\dagger$    &  68.51\%\\
Softmax$^\dagger$ & 66.28\%\\
FC2500$^\dagger$  & 64.64\%\\
FC2500-FC500$^\dagger$ & 65.76\%\\
\hline
DCNN %\citep{atwood2016diffusion} 
	& 70.35\%\\
ChebyNet %\citep{defferrard2016convolutional}
	& 70.92\%\\
GCN  %\citep{kipf2016semi} 
	& 71.01\%\\
MoNet (with GAT) %\citep{monti2016geometric} 
	&  70.60\%\\
MPNN %\citep{gilmer2017neural} 
	& 71.58\% \\
\hline
DSGC (ours) & $\textbf{72.11}\% (\pm 0.285)$ \\
\hline
\end{tabular}
\caption{Accuracy on the validation set of 20NEWS. Results marked with $^\dagger$ come from \cite{defferrard2016convolutional}. The number in the parenthesis is the standard deviation.}
\label{tab:20news}
%\end{table}
\end{wraptable}

Following the experiment in \cite{defferrard2016convolutional}, we test DSGC and other baselines in the text categorization application, and use the 20NEWS dataset (\cite{joachims1996probabilistic}) for our experiments. 20NEWS consists of 18,845 text documents with 20 topic labels.  Individual words in the document vocabulary are the nodes in the graph for convolution. Each node has its word embedding vector generated by Word2Vec algorithm (\cite{mikolov2013distributed}) on the same corpus. Following the experiment settings in \cite{defferrard2016convolutional} we select the top 1000 most frequent words as the nodes. Table \ref{tab:20news} summarizes the results of the graph convolution methods plus three popular traditional classifiers (Linear SVM, Multivariate Naive Bayes and Softmax). DSGC has the best result on this dataset. Notice that the traditional classifiers are trained and tested with the feature set of the top 1000 words, which is the same setting as in the graph convolution models. If all words are used, traditional classifiers would have higher performance.

\begin{comment}
\subsection{DSGC with Multiple Neural Architectures}

\label{sec:advance}

As the proposed DSGC is mathematically compatible with the traditional convolution method by performing channel special filter learning within one-layer, naturally, we can directly replace the convolution layers of general deep convolution frameworks with the DSGC modules while keeping a similar performance without modifying the framework structure.
We examine DSGC with the following frameworks which are popular in recent years for standard convolution over images: 
(1) Inception \citep{szegedy2016rethinking}, 
(2) DenseNet framework \citep{huang2016densely} and 
(3) Squeeze-and-Excitation block \citep{hu2017squeeze}. 
The details of those architectures are included in the Appendix \ref{sec:appendix}. The results are presented in Table \ref{tab:adv}. %and plotted in Figure \ref{fig:advance}. 
Clearly, combined with the advantageous architectures, the performance of DSGC in image classification can be further improved (DSGC-DenseNet over DSGC-VGG-13). It demonstrates that the DSGC can easily enjoy the benefits of framework design for free from the traditional 2d-convolution network community.  

\input{table/advance.tex}
\end{comment}
%\input{src/simulation.tex}

\section{Conclusion}
\label{sec:conclusion}
This paper presents a unified view of graph convolution and grid-based convolution methods, 
and proposes the novel DSGC approach that is applicable to non-grid spatial data. DSGC subsumes several existing graph convolution methods as special cases and is compatible to depthwise separable convolution for image classification by performing channel-special filters learning in data manifold. The proposed DSGC yields state-of-the-art performance on multi-domain benchmark datasets with a relatively small number of model parameters, reasonable computation cost, and is easy to be plugged in different neural network architectures.
For future research we plan to extend DSGC to a broader range of problems, including social network and citation graph analysis,
where the spatial coordinates of the nodes (node embeddings) can be jointly learned along with the convolution filters, or defined by node embedding algorithm.

%\clearpage
\bibliographystyle{apalike}
\bibliography{paper}

%\clearpage
\appendix

\section{Additional Experiment}
\label{sec:appendix-additional}

\subsection{DSGC with Multiple Neural Architectures}

\label{sec:advance}

As the proposed DSGC is mathematically compatible with the traditional convolution method by performing channel special filter learning within one-layer, naturally, we can directly replace the convolution layers of general deep convolution frameworks with the DSGC modules while keeping a similar performance without modifying the framework structure.
We examine DSGC with the following frameworks which are popular in recent years for standard convolution over images: 
(1) Inception \citep{szegedy2016rethinking}, 
(2) DenseNet framework \citep{huang2016densely} and 
(3) Squeeze-and-Excitation block \citep{hu2017squeeze}. 
The details of those architectures are included in the Appendix \ref{sec:appendix-detail}. The results are presented in Table \ref{tab:adv}. %and plotted in Figure \ref{fig:advance}. 
Clearly, combined with the advantageous architectures, the performance of DSGC in image classification can be further improved (DSGC-DenseNet over DSGC-VGG-13). It demonstrates that the DSGC can easily enjoy the benefits of framework design for free from the traditional 2d-convolution network community.  

\begin{table}[!ht]
\centering
\begin{tabular}{l|c|c|c|c|c|c}
\hline
   &\multicolumn{3}{|c|}{Subsampled Images} & \multicolumn{3}{|c}{Original Images} \\
\hline
Method          & CIFAR10 & CIFAR100 & \#params & CIFAR10 & CIFAR100 & \#params \\
\hline
DSGC-VGG-13       
				 & 18.72\% & 44.33\% &5.7M & 7.31\% & 27.29\% &9.9M \\
DSGC-INCEPTION 
				 & 18.27\% & 43.41\% &9.9M & \textbf{6.44}\% & 28.55\% &12M \\
%Our Method-Resnet 
%				 & 18.04\% & 50.01\% &20M & 8.00\% & -\% &20M \\  
DSGC-DenseNet 
				 & 17.17\% & 43.34\% &2.7M & 7.14\% & \textbf{26.50}\% &2.9M \\ 
DSGC-SE 
				 & 18.71\% & 44.15\% &6.1M & 7.00\% & 27.26\% &10M \\ 
\hline
VGG-13           
				 & 18.03\% & 43.42\% &18M & 6.86\% & 26.86\% &18M \\
Xception 
				 & \textbf{17.07}\% & \textbf{41.54}\% & 3.1M & 7.08\% &  26.84\%& 3.1M\\
\hline
\end{tabular}
\caption{Test-set error rates of DSGC-based architectures (first group) and CNNs (second group) %on CIFAR10 and CIFAR100.
}
\label{tab:adv}
\end{table}

\subsection{Training Time Comparison}
\label{sec:traintime}

In Table \ref{tab:traintime}, we report the mean training time per epoch for GCN, DSGC and MoNet. The proposed DSGC computes the convolution weight for each edge in the graph, which requires more computation resources compared to GCN. However, we always perform the graph convolution on a sparse k-nearest neighbor graph, where the number of edges grows only linearly with the node size. Therefore the training is fairly efficient. Notably, DSGC consistently performs better than all graph convolution methods with around 1.5x-4x running time compared to the fastest graph convolution framework (GCN). And MoNet, which also learns multiple filters in a layer but without the channel separation technique applied for DSGC, would be 1x slower than the proposed DSGC method. 

\begin{table}[!ht]
\centering
\begin{tabular}{l|cccc}
\hline
Method           	& CIFAR10 	   	 & TMAX 	  & 20news  \\
\hline
GCN 				& 1.75     & 0.465    & 0.207     \\
MoNet               & 6.87     & 2.81     & 0.550      \\
DSGC (ours)                & 3.81 	   & 1.73     & 0.280 \\
\hline
\end{tabular}
\caption{Training time per epoch for GCN, MoNet and DSGC in three benchmark datasets, meausred in minutes.}
\label{tab:traintime}
\end{table}

\section{Experiment Detail}

\label{sec:appendix-detail}
\subsection{Implementation Details of CIFAR Experiment}

\label{sec:appendix-cifar}

\begin{table*}[!ht]
\centering
\begin{tabular}{c|c|c|c}
\hline
Layers & VGG13 & DSGC-VGG13 & DSGC-DenseNet \\
\hline
Convolution &\conv{2} & \gconv{9}{2} & \gconv{9}{6}\\
\hline
Transition &  & & 1-conv \\
\hline 
Pooling & $2 \times 2$ max-pooling &\multicolumn{2}{|c}{$4$ max-pooling} \\
\hline
Convolution &  \conv{2} & \gconv{9}{2} & \gconv{9}{12}\\
\hline
Transition & & & 1-conv \\
\hline 
Pooling &  $2 \times 2$ max-pooling &\multicolumn{2}{|c}{$4$ max-pooling} \\
\hline
Convolution & \conv{2} & \gconv{9}{2} & \gconv{9}{24}\\
\hline
Transition & & & 1-conv \\
\hline 
Pooling & $2 \times 2$ max-pooling &\multicolumn{2}{|c}{$4$ max-pooling} \\
\hline
Convolution & \conv{2} & \gconv{9}{2} & \gconv{9}{16}\\
\hline
Transition & & & 1-conv \\
\hline 
Pooling & $2 \times 2$ max-pooling &\multicolumn{2}{|c}{$4$ max-pooling} \\
\hline
Convolution & \conv{2} & \gconv{9}{2} & \\
\hline 
Pooling & $2 \times 2$ max-pooling &\multicolumn{2}{|c}{$4$ max-pooling} \\
\hline 
Classifier & \multicolumn{3}{|c}{512D fully-connected, softmax} \\
\hline
\end{tabular}
\caption{Neural Network architecture for CIFAR datasets. Please see the text for more details.}
\label{tab:structure}
\end{table*} 

\begin{figure}[!th]
\centering
\begin{subfigure}{.2\textwidth}
  \includegraphics[width=\linewidth]{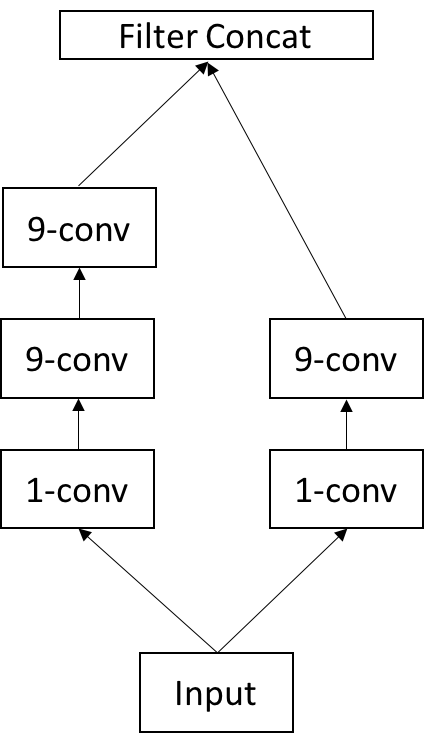}
\end{subfigure}
\caption{Inception Module}
\label{fig:inception}
\end{figure}

In section \ref{sec:cifar} and \ref{sec:advance}, we conduct the experiment on the CIFAR10 and CIFAR100 datasets. We will introduce the architecture settings for the DSGC and baseline models. Table \ref{tab:structure} illustrates the basic architecture used in the experiment. In the DSGC-VGG13 and DSGC-DenseNet models, the $k\text{-conv}$ refers to the spatial convolution (Eq.\ref{eq:proposed}) with $k$-nearest neighbors as the neighbor setting. So the $1\text{-conv}$ is the same as the $1 \times 1 \text{ conv}$, which is doing linear transformation on channels. The hidden dimensions of VGG13 and DSGC-VGG13 are set as $\{256, 512, 512, 512\}$ and $\{256, 512, 512, 1024\}$. The growth rate of DSGC-DenseNet is 32. And the baseline graph and geometric convolution methods use the identical architecture as DSGC-VGG13. For the subsampled CIFAR experiment, We eliminate the first convolution, transition and pooling layer, and change the spatial convolution from 9-conv to \{16-conv, 12-conv, 8-conv, 4-conv\}. For the DSGC-SE, we follow the method described in \cite{hu2017squeeze} to add the SE block to DSGC-VGG13 architecture. We use the dropout scheme described in \cite{huang2016densely} for the DSGC-DenseNet model, and add the dropout layer after the pooling layer for VGG13 and DSGC-VGG13 models. For the DSGC-Inception model, we imitate the design of the Inception Network (\cite{szegedy2016rethinking}). The key idea is letting a convolution layer have different size of convolution filters. We use a simple example as our Inception module, which is illustrated in Figure \ref{fig:inception}. 

For the CNN model, we still format the input signal in the matrix shape. The signals in invalid points are set as 0. Furthermore, to perform the fair comparison with standard CNN in the subsampled situation, we append a mask matrix as an additional channel for input signals to indicate whether the pixel is valid or not. For the ChebyNet, we set the polynomial order as $K = 3$.  

The pooling layer is implemented by K-means clustering. The centroid of each clusters is regarded as the new node after pooling, and its hidden vector is the mean or max over the nodes in that cluster. Notice that, we only normalize the input signals to [0,1] and do not adopt any other data preprocessing or augmentation tricks.

For the $\triangle_{ij}$ used in DSGC and MoNet, we use a 5 dimension feature vector. We denote the coordinate of $i$-th node as $(x_i, y_i)$, and $\triangle x_{ij} = x_i - x_j, \triangle y_{ij} = y_i - y_j, \triangle d_{ij} = \triangle x_{ij}^2 + \triangle y_{ij}^2$. Then $\triangle_{ij} = (sign(\triangle x_{ij}), |\triangle x_{ij}|, sign(\triangle y_{ij}), |\triangle y_{ij}|, \triangle d_{ij})$.

The same learning schedule is applied to all models. We use SGD to train the model for 400 epochs. The initial learning rate is 0.1, and is divided by 10 at 50\% and 75\% of the total number of training epochs.

\subsection{Implementation Details of Time Series Prediction}

\label{sec:appendix-time}
Firstly, we will give the formal definition of the time series forecasting, that is, spatiotemporal regression problem. We formulate the the spatiotemporal regression problem as a multivariate time series forecasting task with the sensors' location as the input. More formally, given a series of time series signals observed from sensors $\bY = \{\by_1, \by_2, \cdots, \by_T\}$ where $\by_t \in \R^n$ and $n$ are the number of sensors, and the locations of sensors $\bL = \{\bl_1, \bl_2, \cdots, \bl_n\}$ where $\bl_i \in \R^2$ and indicates the coordinate of the sensor, the task is to predict a series of future signals in a rolling forecasting fashion. That being said, to predict $\by_{T+h}$ where $h$ is the desirable horizon ahead of the current time stamp $T$, we assume $\{\by_1, \by_2, \cdots, \by_T\}$ are available. Likewise, to predict the signal of the next time stamp $\by_{T+h+1}$, we assume $\{\by_1, \by_2, \cdots, \by_T, \by_{T+1}\}$ are available. In this paper, we follow the setting of the autoregressive model. Define a window size $p$ which is a hyper-parameter firstly. The model input at time stamp $T$ is $\bX_T = \{\by_{T- p +1}, \cdots, \by_{T}\} \in \R^{n \times p}$. In the experiments of this paper, the horizon is always set as 1.

Intuitively, different sensors may have node-level hidden features influencing its propagation patterns and final outputs. For each node, we let the model learn a node embedding vector and concatenate it with the input signals. The embedding size is tuned according to the validation set. By using this trick, each node has limited freedom to interface with its propagation patterns. 

One thing readers may notice is that there are 10\% data in USHCN dataset missing. To deal with that, we add an additional feature channel to indicate which point is missing. For the time series models, we tune the historical window $p$ according to the validation set. For the rest of models, we set the window size $p=18$ for Solar dataset and $p=6$ for USHCN datasets. The network architecture used in this task is 7 convolution layers followed by a regression layer. The $\triangle_{ij}$ setting is the same as the previous one. We use the Adam optimizer \citep{kingma2014adam} for this task, and train each model 200 epochs with learning rate 0.001. 

Except for the graph convolution methods,  we also add in traditional methods of time series forecasting for comparison, such as (1) Autoregressive model (AR) which predicts future signal using a window of historical data based on a linear assumption about temporal dependencies, (2) Vector autoregressive model (VAR) which extends AR to the multivariate version, namely, the input is the signals from all sensors in the history window, and (3) the LSTNet deep neural network model \citep{lai2017modeling} which combines the strengths of CNN, RNN and AR.  None of those methods is capable of leveraging locational dependencies via graph convolution.

\begin{table}[!ht]
\centering
\begin{tabular}{l|c|c|c}
\hline
Method           	& TMAX 	   	 & PRCP 	  & Solar  \\
\hline
AR                	& 8.2354     & 30.3825    & 0.03195        \\
VAR               	& 17.9743    & 29.2597    & 0.03296         \\
LSTNet %\citep{lai2017modeling}          	
					& 10.1973    & 29.0624    & 0.02865   \\
\hline
DCNN %\citep{atwood2016diffusion}
					& 6.5188	 & 29.0424	  & 0.02652	  \\
ChebyNet %\citep{defferrard2016convolutional}
					& 5.5823     & 27.1298    & 0.02531     \\
GCN %\citep{kipf2016semi}
					& 5.4671     & 27.1172    & 0.02512     \\
MoNet (with GAT) %\citep{monti2016geometric}
					& 5.8263     & 26.8076    & 0.02564    \\ 
MPNN 				& 5.3331     & 26.4766	  & 0.02496\\
\hline
DSGC (ours)               & $\textbf{5.1438} (\pm 0.0498)$ & $\textbf{25.8228} (\pm 0.249)$ & $\textbf{0.02453} (\pm 0.00022)$\\
\hline
\end{tabular}
\caption{Test-set performance for graph convolution methods on time series prediction tasks measuring in RMSE. For our method, we report the standard deviation of the performance by running the model with 10 random seeds.}
\label{tab:time-add}
\end{table}

Table \ref{tab:time} summarizes the evaluation results of all the methods, where the performance is measured using the Root Square Mean Error (RMSE). The best result on each dataset is highlighted in boldface. The group of the first three methods does not leverage the spatial or locational information in data. The second group (graph-based convolution methods) consists of the neural network models which leverage the spatial information about sensor networks.  The methods in the second group clearly outperform the methods in the first one, which does not explicitly model the spacial correlation within sensor networks. 

\subsection{Implementation Details of Document Categorization}

The data preprocessing follows the experiment details in \cite{defferrard2016convolutional}. And the network architecture for all models is 5 convolution layers followed by two MLP layers as the classifier. After each convolution layer, a dropout layer is performed with dropout rate of 0.5. The nodes' coordinate is the word embedding, and the method to calculate $\triangle_{ij}$ is similar to the previous ones. The optimizer used in this task is the same as the CIFAR experiment. 

\iffalse
\subsection{Variance of DSGC performance}
\label{sec:variance}

In this section, we report the variance of DSGC method in all 3 tasks. We run the DSGC model for 10 times and report the mean$\pm$std: CIFAR $7.39\pm 0.136$, USHCN-TMAX $5.211\pm 0.0498$, 20news $71.70\pm 0.285$. Obviously, the variance is significantly smaller than the performance gap between the DSGC model and best baseline results (CIFAR 8.34, USHCN-TMAX 5.467, 20news 71.01).
\fi
\end{document}